\documentclass[twoside,11pt]{article}

\usepackage{jmlr2e}
\usepackage{wrapfig}
\usepackage{algorithmic}
\usepackage{amsmath}
\usepackage{multirow}
\usepackage{graphicx} 
\usepackage{subfigure} 
\usepackage{subfigure}
\usepackage{wrapfig}
\usepackage{hyperref}
\usepackage{array}
\usepackage{numcompress}

\ShortHeadings{Prediction of Preterm Birth}{Vovsha et al.}
\firstpageno{1}

\begin{document}

\title{Using Kernel Methods and Model Selection for \\  Prediction of Preterm Birth}

\author{\name Ilia Vovsha, Ansaf Salleb-Aouissi \email {iv2121, as2933}@columbia.edu \\
       \addr Computer Science Department\\
       Columbia University, NY, NY
       \AND
       \name Anita Raja, Thomas Koch, Alex Rybchuk  \email {araja, koch, rybchuk}@cooper.edu \\
       \addr Albert Nerken School of Engineering\\
        The Cooper Union, NY, NY
      \AND
       \name Axinia Radeva,  Ashwath Rajan, Yiwen Huang, Hatim Diab, Ashish Tomar\email {aradeva, anr2121, yh2726, hdiab, ast2124}@columbia.edu \\
       \addr Center for Computational Learning Systems\\
       Columbia University, NY, NY
       \AND
       \name Ronald Wapner \email rw2191@cumc.columbia.edu \\
       \addr Department of Obstetrics and Gynecology\\
       Columbia University Medical Center.
       } 




\maketitle

\begin{abstract}
We describe an application of machine learning to the problem of predicting preterm birth. 
We conduct a secondary analysis on a  clinical trial dataset collected by the National Institute of Child Health and Human Development (NICHD) while focusing our attention on predicting different classes of preterm birth. 
We compare three approaches for deriving predictive models: 
a support vector machine (SVM) approach with linear and non-linear kernels, logistic regression with different model selection along with a model based on decision rules prescribed by physician experts  for prediction of preterm birth.
Our approach highlights the pre-processing methods  applied  to handle the inherent dynamics, noise and gaps in the data  and describe techniques used to handle skewed class distributions.
Empirical experiments demonstrate significant improvement in predicting preterm birth compared to past work. 

\end{abstract}

\section{Introduction}
\label{intro}

%
%

{P}{remature} or preterm birth (PTB) is   a major long-lasting public health problem with heavy emotional and financial consequences to families and society \citep{born_too_soon,conova2016}. PTB is the leading cause of neonatal mortality and, long-term disabilities. Furthermore, over 26 billion dollars are spent annually on the delivery and care of the 12-13\% of infants who are born preterm in the United States  \citep{behrman2007preterm}.  A crucial challenge is to identify women who are at the highest risk for very early preterm birth and to develop interventions.  Equally important, would be the ability to identify women at the lowest risk to avoid unnecessary and costly interventions.
A particularly challenging population to determine PTB risk is first time mothers (nulliparous women) due to the lack of prior pregnancy history. 

Prediction of preterm birth represents a compelling application from a machine learning perspective. 
It has been an exceedingly challenging problem, predominantly due to (1) the inherent complexity of its heterogeneous multifactorial etiology, (2) the temporal dynamics of pregnancy and (3) the lack of approaches capable of integrating and interpreting large multidimensional data.


 
Risk factors of PTB are heterogenous and include history of PTB, race, age, parity of the mother,  bacterial vaginosis, urinary tract infection, smoking, bleeding, cervix length.
Most studies to date have examined individual risk factors independently of each other through univariate analyses of  their coincidence with PTB.
While these studies led to many insights on the PTB problem, current models lack sufficiently good prediction to be used clinically \citep{Mercer19961885}.
Previous results on this dataset using a multivariate logistic regression model show a sensitivity of 24.2\% and 18.2\%, and  specificity of 28.6\% and 33.3\%, for nulliparous and multiparous women respectively. 

We describe our efforts towards developing multivariate linear and non-linear models that integrate all risk factors for predicting preterm birth.\footnote{A preliminary version of this work appeared in ~\citep{SSS147694}.}
We use the ``Preterm Prediction Study,'' a clinical trial dataset collected by the National Institute of Child Health and Human Development (NICHD) -- Maternal-Fetal Medicine Units Network (MFMU).  
We compare three approaches for deriving predictive models: 
a support vector machine (SVM) approach with linear and non-linear kernels, logistic regression with model selection along with a hand-picked model. 

We also focus our attention (as recommended by \citep{nichd2010}) on predicting (1) any kind of preterm birth, (2) spontaneous preterm birth, and (3) predicting preterm birth for nulliparous women. 
Furthermore, etiologies of preterm birth are believed to be different as pregnancy progresses. Hence, we also derive models at different time points, which represent the three main prenatal visits in the preterm prediction study, that is at 24, 26 and 28 weeks gestation. 
Our results for  the spontaneous preterm birth class at 28 weeks gestation show an improvement of 20\% and 30\% for sensitivity and specificity respectively as compared to \citep{Mercer19961885}. In addition, we obtain approximately $50\%$ sensitivity and specificity across other data classes and time points. 

This paper is  organized as follows: in Section \ref{background}, 
we provide an overview of the risk factors and state-of-the-art systems for predicting PTB. 
We describe the Preterm Prediction Study dataset in Section \ref{mfmu} along with our pre-processing methods. 
We present our approach in Section \ref{approach} and our empirical evaluation along with a discussion of our results in Section \ref{experiments}. 
Finally, we discuss the significance and impact of this study and outline future work in Section \ref{future}.

\section{Background}
\label{background}

%
%
In this section, we describe the known risk factors for PTB
and review state-of-the-art approaches to devise a risk-scoring system for PTB.

\paragraph{Risk Factors for Preterm Birth:}
Approximately 30\% of preterm deliveries are indicated based on maternal or fetal conditions such as mother's preeclampsia and intra uterine growth restriction. 
The remaining 70\%, known as spontaneous PTB (SPTB), 
occur following the onset of spontaneous preterm labor, prelabor Premature Rupture Of the Membranes (pPROM), or cervical insufficiency \citep{Goldenberg200875}. 
Spontaneous preterm labor is a heterogeneous condition, the final common product of numerous biologic pathways that include immune, inflammatory, neuroendocrine, and vascular processes \citep{behrman2007preterm}. 
Epidemiological investigations 
have largely associated single factors with PTB.
Of the many risk factors for preterm labor, 
a prior history of preterm delivery is the most predictive with a  
recurrence risk as high as 50\% depending on the number and gestational age of previous deliveries \citep{Goldenberg200875}. 
It has been shown~\citep{goldenberg1998} that the odds ratio of SPTB was highest for a positive fetal fibronectin test, followed by short cervix ~\citep{crane2008tra} and history of prior PTB. 
However, in practice, prior history of preterm delivery is  used as the most predictive indicator of PTB in most  clinical settings.
Risk factors include race \citep{Goldenberg200875}, low socioeconomic status, extremes in age, 
single marital status \citep{Smith01012007,PPE:PPE711}, 
low pre-pregnancy body mass index, \citep{Hendler2005882}, 
and high-risk behaviors during pregnancy (e.g. tobacco, cocaine and heroin use). 
Psychological factors \citep{NancyK.Grote10012010}, \citep{Zhu201034e1}, and obstetrical conditions \citep{GHA,Gotsch01111,BJO:BJO1120,Tita2010339} are also known to increase the risk of PTB. 
Additional risk factors include closely spaced gestations \citep{Conde-Agudelo02222}, multiple gestations \citep{Goldenberg200875}, 
assisted reproductive technologies \citep{Allen04444},  exposure to tobacco smoke  \citep{Kharrazi05555,Jaakkola2001}, 
and genetic factors \citep{Porter01515,Winkvist01041998}.
Those judged at risk for PTB are typically treated by prenatal administration of progesterone  17 OHPC (IM progesterone) \citep{Acog_2003,Flood2012}. However, pregnant nulliparous women are often not treated due to the lack of prior pregnancy history.

\paragraph{Risk Scoring Systems for Predicting Preterm Birth:} 
\label{riskscoring}
In the late 1960's, Papiernik proposed an empirical method for estimating the risk of premature delivery 
\citep{papiernik-berkhauer1969}.  In this approach, maternal characteristics are grouped into four series of comparable variables (social status, obstetric history, work conditions, pregnancy characteristics) in a two-dimensional table. 
Point values varying from 1 to 5 according to the degree of their importance are assigned to all characteristics. 
The sum of the points gives the risk of Premature delivery. 
Papiernik's risk table was later modified by Creasy et al. and used in the risk of preterm delivery (RPD) system proposed in \citep{creasy1980} (Appendix, Table \ref{table:creasy}). 
Further assessment of the prediction performance of Creasy's table \citep{Edenfield1995} on another population has shown low performance.
 
Another graded risk system was proposed \citep{Mercer19961885} in the context of the NICHD MFMU preterm prediction study. The results of a multivariate 
logistic regression were modest  with sensitivity of 24.2\% and 18.2\%; specificity of 28.6\% and 33.3\%, respectively for nulliparous and multiparous women. 
This constitutes our baseline for comparison.

\citet{goodwindat} have explored the use of data mining techniques to predict preterm labor. 
They have identified seven demographic variables that predict preterm birth.
While these results are interesting, there are concerns whether the sampling of a particular demographic  would be representative of more general population. 
Furthermore, the experiment procedure is unclear -- for example, the Area Under Curve (AUC) could have been obtained on a validation set or an unseen test set; consequently it is difficult to reproduce their results.
\citet{CourtneySPG08} describe a secondary analysis showing that the demographic preterm prediction model generated in \citep{goodwindat} 
generalizes to a broader population with a modest accuracy. 
Today, there is no widely tested risk scoring/prediction system that combines PTB factors \citep{davey2011ris}. 

Our models differ from the previous work as follows:
(1) the dataset we study represents a diverse population from ten medical centers across the US, 
(2) we derive predictive models at different stages in pregnancy,
(3) we derive models for specific classes of patients, namely nulliparous women and also for spontaneous PTB, and
(4) the procedure we use to evaluate our models is robust and reproducible.



\section{The Preterm Prediction Study Data}
\label{mfmu}
\begin{figure*}[htbp]
\begin{center}  \includegraphics[scale=0.7]{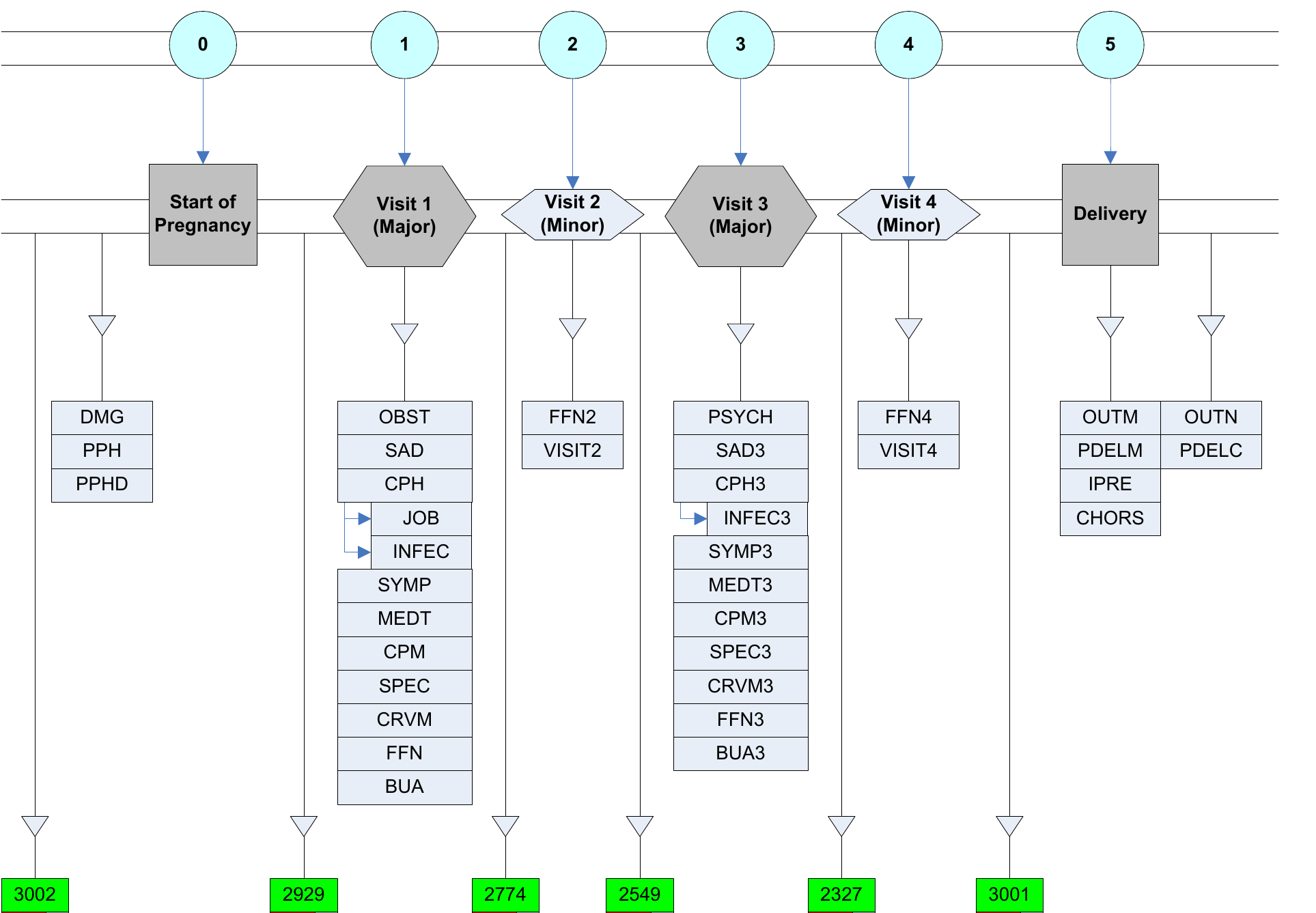}
{\scriptsize
\begin{tabular}{|p{1.05cm}l|p{1.05cm}l|}
\hline
{\bf Acronym}	&	{\bf Description}	&	{\bf Acronym}	&	{\bf Description}	\\ \hline 
DMG	&	 Demographics and Home Life	&	CRVM	&	 Cervical Measurements V1	\\
PPH	&	 Previous Pregnancy History	&	CRVM3	&	 Cervical Measurements V3	\\
PPHD	&	 Previous Pregnancy History Detail 	&	FFN	&	 Fetal Fibronectin Analysis V1	\\
OBST	&	 Obstetrical  and Medical Complications 	&	FFN2	&	 ÊFetal Fibronectin Analysis V2 	\\
SAD	&	 Substance Use V1	&	FFN3	&	 ÊFetal Fibronectin Analysis V3 	\\
SAD3	&	 Substance Use V3	&	FFN4	&	 ÊFetal Fibronectin Analysis V4 	\\
CPH	&	 Current Pregnancy History V1	&	BUA	&	 Blood and Urine Analysis V1 	\\
CPH3	&	 Current Pregnancy History V3	&	BUA3	&	 Blood and Urine Analysis V3 	\\
JOB	&	 Current or Last Job	&	PSYCH	&	 Psychological Questionnaire 	\\
INFEC	&	 Infections During This Pregnancy V1	&	VISIT2	&	 Yeast  and Intercourse Variables V2 	\\
INFEC3	&	 Infections During This Pregnancy V3 	&	VISIT4	&	 Yeast and  Intercourse Variables V4 	\\
MEDT	&	 Medications and Treatments V1 	&	OUTM	&	 ÊPregnancy Outcome, Maternal Data 	\\
MEDT3	&	 Medications and Treatments V3 	&	OUTN	&	 ÊPregnancy Outcome, Neonatal Data 	\\
SYMP	&	 Symptoms During Previous Week V1 	&	OUTS	&	 Pregnancy Outcome Status	\\
SYMP3	&	 Symptoms During Previous Week V3 	&	IPRE	&	 Indicated Preterm Birth Reasons 	\\
CPM	&	 Current Pregnancy Measurements V1 	&	CHORS	&	 Chorioamnionitis Suspected 	\\
CPM3	&	 Current Pregnancy Measurements V3 	&	PDELM	&	 Preterm Delivery, Maternal Data 	\\
SPEC	&	 Specimen Collection V1	&	PDELC	&	 Preterm Delivery, Clinical Data 	\\
SPEC3	&	 Specimen Collection V3	&	~~	&		~~\\ \hline 
 \end{tabular}
 }
 
\caption{Illustration of MFMU data timeline and description of the set of feature groups.
The numbers at the bottom of the diagram indicate the number of patients that reached that point in time of the study. 
These numbers decrease with time for several possible reasons including: patients withdrawing from study/delivered/lost to follow up/skipped major visit etc. 
The last number (3001) indicates the total number of patients with known 
pregnancy outcomes. }
\label{timeline}
\end{center}

\end{figure*}

%
%
We have obtained the released data set for the {\em Preterm Prediction Study}, performed by the NICHD Maternal Fetal Medicine Units (MFMU) Network between 1992 to 1994.
This study is an observational prospective study of 3,073 women with singleton pregnancies recruited at less than 24 weeks  gestational age. 
Of the women enrolled, 2,929 participated in the study at the 10 participating MFMU centers across the United States between October 1992 and July 1994. 
There were 1,711 multiparous and 1,218 nulliparous women. 
The incidence of spontaneous preterm birth was 10.3\% overall 8.2\% for nulliparous and 11.9\% for multiparous women \citep{Mercer19961885}. 
Henceforth, we will refer to this data as the {\em MFMU data}.
Participating women in this study were followed up by research nurses during four visits at 24 (time T0), 26 (time T1), 28 (time T3) and 30 (time T4) weeks gestation for screening tests. 
The MFMU data timeline is illustrated in Figure \ref{timeline}.
The data collected from all visits  \cite{mfmu_moo} has over 400 variables in all. These include demographic, behavioral, medical history, previous and current pregnancy history, digital cervical examination, vaginal ultrasound, cervical and vaginal fetal fibronectin, KOH prep for yeast tests, and a psychosocial questionnaire. 
The detailed outcome of the pregnancies is as follows for spontaneous PTB $<$32 weeks (2\%),  $< $35 weeks (4\%), $<$37 weeks (10\%); indicated PTB $<$37 weeks (4\%).
The Preterm prediction cohort singletons was released only in April 2007 under the study title: {\em  Screening for Risk Factors for Spontaneous Preterm Delivery in Singletons and Twins} \citep{MFMU}. 

%
%
%

\begin{table}[h]
\begin{center}
\vspace*{-0.2in}
\caption{(Left) \texttt{CPM} group features; (Bottom) \texttt{Right} group features. From left to right: feature number in the group, name, feature number in the raw data, type, range, number of missing values, processing flag, description.   }
\includegraphics[scale=0.75]{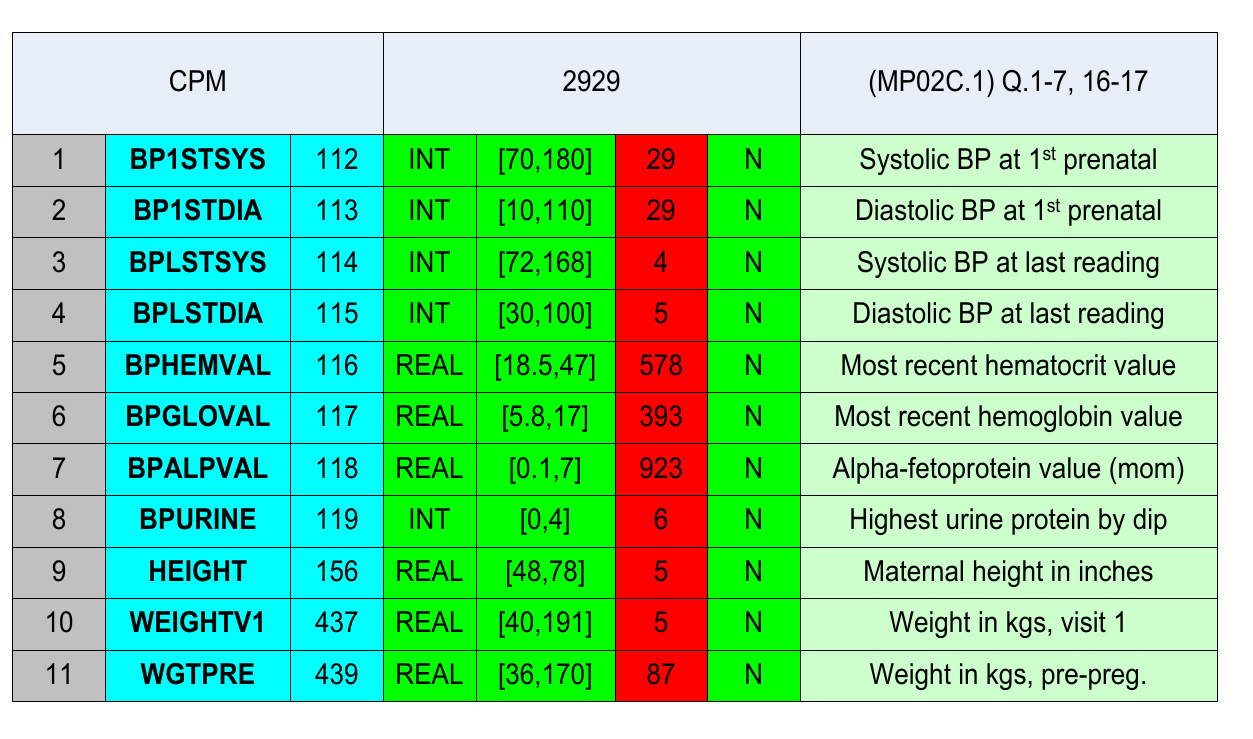} \hspace*{0cm} \includegraphics[scale=0.75]{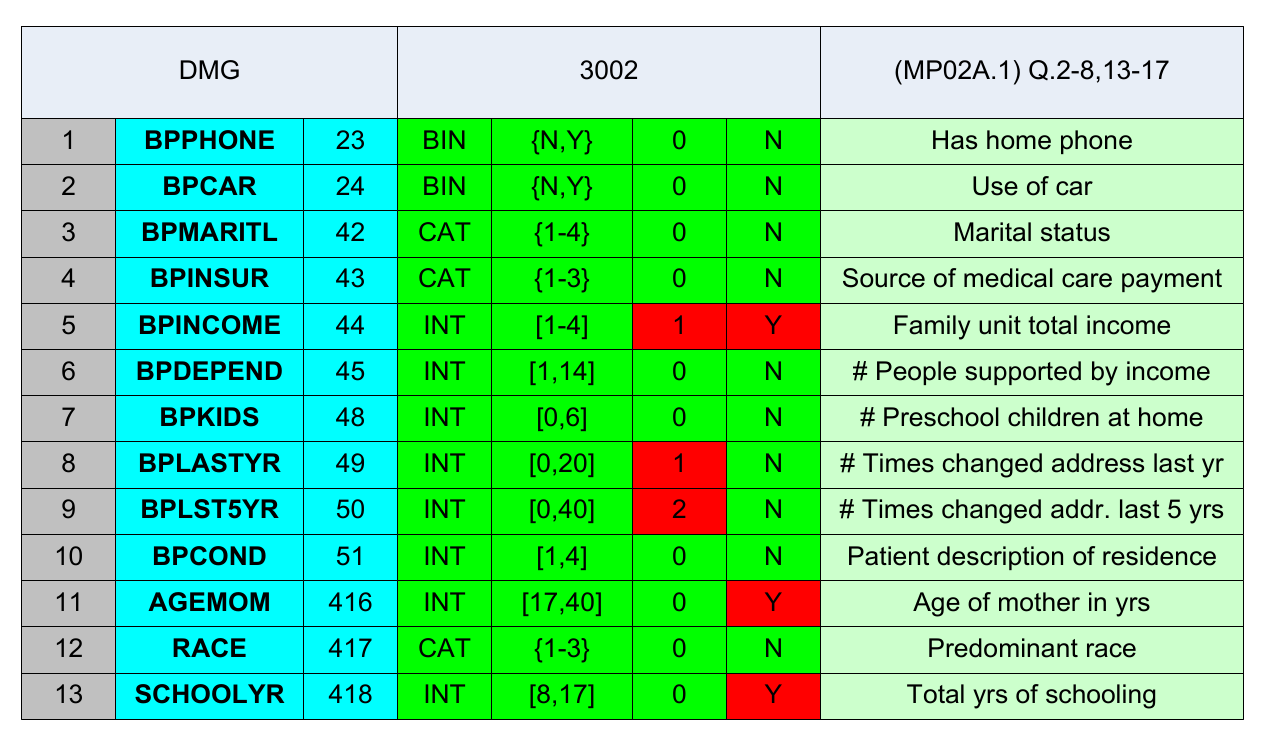}
\label{CPM_DMG}
\vspace*{-0.3in}
\end{center}
\end{table}

The MFMU data is a very rich and highly structured dataset.
As a result, multiple processing steps are required.
We face several challenges,  including the complexity of data, missing data and skewed class distribution (addressed in Section 4). 
For reproducibility purposes, we describe our preprocessing steps in considerable detail in  \cite{mfmu_cing}.

%
%
%


\paragraph{Complexity of Data:}

We handle the complexity of the data by organizing features
into groups (according to the original questionnaire)
as depicted in Figure \ref{timeline}.
At each visit, a set of feature groups is collected.
We focus our study on the three major visits at time T0, T1 and T3.

%

Since features are obtained from various sources, they are not always uniform.
We undertake several processing steps to convert the data into a standard numerical format 
suitable for off-the-shelf machine learning algorithms.
In particular, Yes/No features are converted to binary (1/0) values, 
categorical features are converted to a set of binary features,
unusual values (e.g., ``$>3$'', ``2-3'') are replaced with reasonable approximations (4, 2.5 respectively),
and features with arbitrary ranges are normalized to the [0,1] interval.
We review each feature with non-standard values manually and decide what is the 
most appropriate processing step. 

For example, consider some of the features from the ``\texttt{DMG}'' group 
(Table \ref{CPM_DMG}).
The \texttt{BPPHONE} (has home phone) and \texttt{BPCAR} (use of car) are Yes/No features,
\texttt{BPMARITL} (marital status) is a categorical feature with four categories,
and the \texttt{AGEMOM} (age of mother) and \texttt{SCHOOLYR} (total years of schooling) 
features both have unusual values and different integer ranges.
We replace unusual values in the the last two features (\texttt{AGEMOM, SCHOOLYR}) 
by rounding off from below and from above, e.g., for \texttt{AGEMOM}, ``$\leq 17$'' is replaced with 17
and ``$\geq 40$'' is replaced with 40.

%
%

Furthermore, PTB data is characterized by complex interdependencies among its features (physiological as well as socio-economic) which contributes to the difficulty of accurate prediction of PTB. We propose that use of non-linear methods like using the RBF kernel would pick up on these complex non-linear interdependencies and improve the prediction accuracy. We also propose to use logistic regression with model selection to automatically include whole groups of co-advisorlinear predictors and hence take into account this aspect of the data. 

\paragraph{Missing Data:}


Our main objective in this work is to retain as many features and examples as possible.
Hence, we prefer to fill in (complete) values rather than delete features. Since a substantial number of features is missing, we follow a simplified approach
and treat features equally whether they are randomly or structurally absent.
Most missing values can be reasonably completed by inserting a default value (e.g., `No' or 0), 
the most common value (for categorical features), or the mean value (for numeric features).
However, some features require non-trivial processing steps, and for those we sometimes include
the range, mean or median, and other features in the computation as well.

As a concrete example, consider the features from the ``\texttt{INFEC}" group. 
All 10 features in this group are Yes/No features that can be structurally absent if
the patient did not report any infections during pregnancy. 
As such, we complete any missing values with the default `No' value.
On the other hand, the features from the ``\texttt{CPM}'' group are all numbers from some range of values (Table \ref{CPM_DMG}). 
All 11 of these features can be randomly missing due to the patient 
not undertaking a test or measurement not being reported after a visit.
For several of these features (1-4), we prefer to complete 
the missing values with the mean of the actual responses. 
However, other features (5-7) have too many missing values (as shown in Table \ref{CPM_DMG}, column 6), 
and so we (reluctantly) remove these features from the dataset.
Finally, some features (9-11) have particular meaning (weight, height at different points
in the pregnancy), and hence we apply non-trivial processing steps
to impute the value from the available information.
For example, if the weight is not measured during visit 1, then
we set feature 10 (\texttt{WEIGHTV1}) to the weight before pregnancy (feature 11)
plus the average difference between the weights of all mothers at visit 1  
and their weights before pregnancy.

\section{Method for Prediction of PTB}
\label{approach}
%
%

In this section, 
 we consider support vector machines (SVMs) and regression methods with model selection.
%
%
We frame PTB as a binary classification problem,
where patients who deliver a baby preterm (full-term) are assigned the positive (negative) class respectively. 
At every tick $(0,1,3)$, each patient (example) is described by a complete feature
vector (see Section~\ref{mfmu}) and a label $(x_i,y_i)$, $y_i\in \{+1,-1\}$.


To validate our results, we repeat the following procedure throughout: 
each dataset is randomly divided into train and test sets with an 80/20 ratio, 
and each class is split proportionally between the sets.
We then apply 5-fold cross-validation (CV) to the train set 
to determine the best model and optimal parameters (if any). 
The best model is tested on the (unseen) test set, and
confusion matrices for various subsets of the data are recorded. 

 The metrics used are sensitivity (the percent of positive instances that are correctly predicted as positive), specificity (the percent of negative instances  that are correctly predicted as negative) and the geometric mean (the square root of the product of the specificity and sensitivity). 
%
%
\paragraph{Support Vector Machines}

We use support vector machines (SVMs) that belong to the family of maximum margin classifiers \citep{VapnikBook}.  
%
The standard approach is to solve the soft-margin formulation \citep{Boser92,Cortes95}:
\begin{eqnarray*}
\min_{w,\xi} && \frac{1}{2}||w||^2 + C\sum_{y_i} \xi_i \nonumber\\
\mathrm{s.t.} &&  y_k[w^\top x_k + b] \ge 1-\xi_k, \:\:\:\ \xi_k \geq 0 \:\:\:\  \forall k \in 1,...,n
\end{eqnarray*}
where $C$ is a positive constant determining  the tradeoff between maximizing the margin and minimizing the misclassifications.  
The $\xi$'s are slack variables that allow to calculate the misclassifications. An example $x_i$ is misclassified if its corresponding slack variable $\xi_i\geq 1$. A margin error occurs if  $0 \leq \xi_i\leq 1$. 
A large C corresponds to assigning a higher penalty to errors. This is useful, since in practice, data is rarely linearly separable.
Typically, the SVM produces a classifier that labels examples $x$ with $y = sign(w^T. x+b)$. 

To account for the large discrepancy between the number of examples in each class,
we scale the hinge loss penalty from the cost function proportionally to the size of each class.
The cost function is thus a slightly modified version of the soft-margin SVM formulation: 
\begin{eqnarray*}
\min_{w,\xi} && \frac{1}{2}||w||^2 + C_{-}\sum_{y_i=-1} \xi_i + C_{+}\sum_{y_j=+1} \xi_j\nonumber\\
\mathrm{s.t.} &&  y_k[w^\top x_k + b] \ge 1-\xi_k,  \:\:\:\ \xi_k \geq 0 \:\:\:\ \forall k \in 1,...,n
\end{eqnarray*}
%
By assigning different misclassification costs, we can give
equal overall weight to each class in measuring performance. 
In order to avoid tuning two cost parameters, we set: \\
$ C_{+} \times n_{+} = C_{-}\times n_{-} $ where $n_{+}$($n_{-}$) is the number of 
positive (negative) examples \citep{Weston10}.
In our experiments, we use an SVM  with linear, polynomial of degree 2 and 3 along with a radial basis function (RBF) kernels.




%
%
\paragraph{Logistic and Lasso Regression:}

%

The regression study in this paper is motivated by the desire to create a meaningful baseline model 
to evaluate the performance of linear models in this problem space and assess the benefit of model selection methodologies.  
We consider two logistic regression model selection methodologies:  
$l_1$ lasso regression and elastic net regression 
\citep{Zou05regularizationand,tibshirani96regression}. 
Lasso regression uses an ($l_1$ norm) penalty to encourage sparse solutions and perform a level of feature selection. 
Elastic net regression combines the sparsity induction of the $l_1$ norm 
to eliminate the trivial covariates, while using the ridge regression $l_2$ norm 
to automatically include whole groups of collinear predictors once a single covariate 
is added \citep{Zou05regularizationand}.

%

Since the class distribution is skewed towards the negative (full term) class (skewed distribution challenge from the previous section),
we use oversampling techniques to achieve 1:1 levels of negative to positive examples.
Specifically, we use the adaptive synthetic (ADASYN) sampling approach \citep{ADASYN} to adaptively generating minority data to balance the dataset. 
\paragraph{The Creasy Baseline:}
We implemented the risk of preterm delivery score proposed by Creasy, as discussed in Section~\ref{riskscoring}, in order to provide a baseline for comparison. In Creasy's system, patients were initially screened, and then given a follow up screening at 26 to 28 weeks' gestation. This follow up screening provided additional information on the pregnancy and improved the accuracy of the risk of preterm delivery (RPD) system \citep{creasy1980}. Similarly, our implementation had an initial screening at T0 and then added information at the T1 and T3 time points. Our implementation had two key differences to the RPD system. 

In RPD, patients are given a risk score based on factors involving socioeconomic status, past history, daily habits and status of the pregnancy. Consider for instance a mother who encountered DES (diethylstilbestrol) exposure, has a very low socioeconomic status, and is diagnosed with hypertension. This patient is assigned a score of 5 in our implementation. We manually classified  these factors into three feature categories. In the first category, features did not require any modification and are used as is, such as having two children at home or hypertension. The second category of  factors were simply discounted because they were either no longer prevalent (DES exposure) or the dataset lacked information on that feature (head being engaged). In the third category, features were ambiguous. For example, Creasy does not provide a clear distinction between low socioeconomic status and very low socioeconomic status. In order to compensate, we assigned a reasonable definition so that their score could be taken into account.  The RPD scoring system as well as the mapping we developed of RPD risk score factors to MFMU features are available in the Appendix (Tables \ref{table:creasy}, and \ref{RPDmapping1}).


In RPD, the numerical risk factor is then translated into low, medium, or high risk of PTB, whereas SVM and regression methodologies are marked as only low or high risk. We modified RPD by eliminating the medium category. We ran the analysis twice, using seven and thirteen points as our boundary between low and high risk (respectively marked Creasy-7 and Creasy-13 in Table \ref{tables}). A cut-off of seven points agreed with Creasy's definition of medium risk, but did not produce a similar distribution of PTB patients found in \citep{creasy1980}, whereas a cut-off of thirteen produced a similar distribution.

\section{Empirical Evaluation}
\label{experiments}
\renewcommand{\arraystretch}{0.9}

\paragraph{Results:}

%
%

For each of the three problems above, 
we derive prediction models at different time points (ticks). 
Each tick (T0, T1, T3) represents a critical point (major visit) 
at which information is collected. 
In Table~\ref{table_distr}, we list the ratio of positive to negative examples 
in each dataset and the number of features at each tick.

We use the glmnet package to run our regression experiments.
The package implements coordinate descent to train
the elastic net and lasso models \citep{Friedman2010}. We display results for models with weighting factor $r=1$ only, 
as there is little difference in performance between using $r=2$ and $r=1$. We use the sklearn package for random forest and used the "class weight" parameter to handle the data imbalance. We obtain our SVM models with modified code based on the LIBSVM package \citep{LIBSVM}.

\begin{table*}[h]
{\footnotesize

\begin{center}
\caption{Class size and feature count \label{table_distr}}{
 \begin{tabular}{ l | c c c}
  & {\bf T0} & {\bf T1} & {\bf T3}  \\
  \hline
  {\bf Feature count }& 50 & 205 & 316 \\ 
  \hline 
  All data  & 434 / 2,568 & 423 / 2,506 & 334 / 2,215 \\ 
  Spontaneous only & 309 / 2,568 & 302 / 2,506 & 240 / 2,215 \\ 
  Nulliparous only & 156 / 1,087 & 153 / 1,065 & 112 / 951 \\
 \hline
 \end{tabular}}
\end{center}
}
\vspace*{-0.3in}
\end{table*}

We present our results in Table~\ref{tables}.
For each algorithm, we show the sensitivity, specificity, 
and geometric mean (g-mean) performance measures (rounded off to two decimal places) 
on the unseen test set, averaged over five runs plus-minus the standard deviation.

\begin{table*}[h]
{\footnotesize
\begin{center}
\caption{Average test rates for all algorithms at each tick\label{tables}}{

 \begin{tabular}{ l | 
   >{\centering\arraybackslash}p{0.65cm}
   >{\centering\arraybackslash}p{0.65cm} >{\centering\arraybackslash}p{0.7cm} |
   >{\centering\arraybackslash}p{0.7cm} 
   >{\centering\arraybackslash}p{0.7cm} >{\centering\arraybackslash}p{0.7cm} | 
   c c c}      \hline
  &  \multicolumn{3}{c|}{\it{\bf Sensitivity}} & 
     \multicolumn{3}{|c|}{\it{\bf Specificity}} &
     \multicolumn{3}{|c}{\it{\bf g-mean}} \\ 
  \hline
  & {\bf  T0} & {\bf T1} & {\bf T3} & {\bf T0} & {\bf T1} & {\bf T3} & {\bf T0} & {\bf T1} & {\bf T3}  \\ 
  \hline
  \noalign{\medskip} 
  \multicolumn{10}{c}{\bf{Preterm vs. Fullterm, All data}} \\
  \noalign{\medskip}
  \hline
  Lasso	& 0.59 & 0.52 &	0.50 & 0.59 & 0.67 & 0.73 
  & 0.59 $\pm$ 0.02 & 0.59 $\pm$ 0.02 & 0.60 $\pm$ 0.03 \\
  Elastic Net &	0.59 & 0.51 & 0.50 & 0.59 & 0.67 & 0.73
  & 0.59 $\pm$ 0.02 & 0.59 $\pm$ 0.02 & 0.60 $\pm$ 0.03 \\
  Linear SVM  & 0.40 & 0.43 & 0.45 & 0.83 & 0.82 & 0.84 
  & 0.58 $\pm$ 0.04 & 0.59 $\pm$ 0.02 & 0.62 $\pm$ 0.03 \\
Poly. SVM d2&  0.56  &   0.62 &    0.67 &   0.63   &  0.65   &  0.66   
& 0.59  $\pm$  0.03&    0.64   $\pm$ 0.01 &  0.6 $\pm$ 0.04 \\
Poly. SVM d3&  0.55 &     0.27 &     0.23   &  0.62 &     0.87 &   0.93
& 0.57  $\pm$ 0.03 &    0.49   $\pm$ 0.02 &  0.46 $\pm$ 0.07 \\
 RBF SVM & 0.58 & 0.55 & 0.59 & 0.62 & 0.72 & 0.72 
  & 0.60 $\pm$ 0.04 & 0.63 $\pm$ 0.03 & 0.65 $\pm$ 0.03 \\
 Random Forest  & 0.3 & 0.3 & 0.3 & 0.86 &0.88 & 0.93 
  & 0.51 $\pm$ 0.07 & 0.48 $\pm$ 0.09 &0.5 $\pm$ 0.1\\
  Creasy-7 & 0.30 & 0.22 & 0.21 & 0.88 & 0.91 & 0.93 &
  0.52 $\pm$ 0.04 & 0.45 $\pm$ 0.03 & 0.44 $\pm$ 0.03 \\
  Creasy-13 & 0.29 & 0.31 & 0.31 & 0.86 & 0.89 
  & 0.91 & 0.50 $\pm$ 0.04 & 0.52 $\pm$ 0.06 & 0.53 $\pm$ 0.03 \\
  \hline

  \noalign{\medskip}
  \multicolumn{10}{c}{\bf{Preterm vs. Fullterm, Spontaneous only}} \\
  \noalign{\medskip}
  \hline
  Lasso	& 0.53 & 0.35 & 0.36 &	0.54 &	0.66 & 0.67 
  & 0.53 $\pm$ 0.02 & 0.48 $\pm$ 0.04 & 0.48 $\pm$ 0.05 \\
  Elastic Net	& 0.52 & 0.36 &	0.36 & 0.55 & 0.65 & 0.67 
  & 0.53 $\pm$ 0.03 & 0.48 $\pm$ 0.04 & 0.49 $\pm$ 0.05 \\
  Linear SVM & 0.50 & 0.53 & 0.47 & 0.50 & 0.51 & 0.57 
  & 0.49 $\pm$ 0.02 & 0.52 $\pm$ 0.03 & 0.52 $\pm$ 0.02 \\
Poly. SVM d2&  0.56    & 0.44   & 0.41    & 0.48   &  0.58   & 0.6
& 0.51  $\pm$ 0.03 &    0.5   $\pm$ 0.02 &  0.49 $\pm$ 0.03 \\
Poly. SVM d3&  0.42  &  0.17  &   0.02 &   0.62  &   0.86  &   0.93   
& 0.51  $\pm$  0.01 &    0.38   $\pm$ 0.08 &  0.11 $\pm$ 0.1 \\
 RBF SVM  &  0.40 & 0.40 & 0.43 & 0.59 & 0.60 & 0.58  
  & 0.49 $\pm$ 0.04 & 0.49 $\pm$ 0.04 & 0.50 $\pm$ 0.02 \\
 Random Forest  & 0.08 & 0.03 & 0.03 & 0.95 &0.97 & 0.98 
  & 0.2 $\pm$ 0.1 & 0.13 $\pm$ 0.1 &0.14 $\pm$ 0.1\\
  Creasy-7 & 0.09 & 0.10 & 0.10 & 0.88 & 0.89 & 0.92 
  & 0.28 $\pm$ 0.03 & 0.30 $\pm$ 0.02 & 0.30 $\pm$ 0.03 \\
  Creasy-13 & 0.07 & 0.11 & 0.08 & 0.88 & 0.90 & 0.91 
  & 0.25 $\pm$ 0.08 & 0.30 $\pm$ 0.05 & 0.26 $\pm$ 0.07 \\

  \hline 		

  \noalign{\medskip}
  \multicolumn{10}{c}{\bf{Preterm vs. Fullterm, Nulliparous only}} \\
  \noalign{\medskip}
  \hline
  Lasso & 0.36 & 0.35 &	0.31 &	0.58 & 0.68 & 0.75 
  & 0.46 $\pm$ 0.06 & 0.48 $\pm$ 0.05 & 0.47 $\pm$ 0.11 \\
  Elastic Net &	0.35 &	0.35 &	0.30 &	0.58 &	0.69 &	0.76 
  & 0.45 $\pm$ 0.07 & 0.49 $\pm$ 0.06 & 0.47 $\pm$ 0.06 \\
  Linear SVM  & 0.40 &  0.40 &  0.42 &  0.59 &  0.60 &  0.66 
  &  0.48 $\pm$ 0.03 &  0.49 $\pm$ 0.06 &  0.52 $\pm$ 0.07 \\
Poly. SVM d2&  0.49  &   0.38   &  0.38    & 0.46    &  0.63   &  0.73
& 0.46  $\pm$  0.04&    0.48   $\pm$ 0.04 &  0.52 $\pm$ 0.05 \\
 Poly. SVM d3&  0.38    & 0.15   &  0.18    & 0.61    & 0.88   &  0.93
& 0.48  $\pm$ 0.05 &    0.36   $\pm$ 0.05 &  0.4 $\pm$ 0.06 \\
  RBF SVM   & 0.41 &  0.34 &  0.42 &  0.64 &  0.64 &  0.68 
  &  0.50 $\pm$ 0.05 & 0.46 $\pm$ 0.06 & 0.53 $\pm$ 0.08 \\
 Random Forest  & 0.12 & 0.09 & 0.03 & 0.92 & 0.93 & 0.98 
  & 0.31 $\pm$ 0.09 & 0.25 $\pm$ 0.14 & 0.14 $\pm$ 0.1\\
 Creasy-7 & 0.02 & 0.13 & 0.13 & 0.87 & 0.88 & 0.92 
  & 0.06 $\pm$ 0.13 & 0.34 $\pm$ 0.05 & 0.35 $\pm$ 0.05 \\
  Creasy-13 & N/A & 0.10 & 0.14 & N/A & 0.88 & 0.90 
  & N/A $\pm$ N/A & 0.22 $\pm$ 0.21 & 0.27 $\pm$ 0.25 \\
 \hline

 \end{tabular}}
\end{center}
}
\end{table*}

%

\paragraph{Observations:}


As we stated in Section \ref{background}, the test error of a multivariate logistic  regression model was modest \citep{Mercer19961885} with sensitivity of 24.2\% and 18.2\%; specificity of 28.6\% and 33.3\%, respectively for nulliparous and multiparous women.  This constitutes our baseline for comparison. 
Our results for the spontaneous preterm birth class at 28 weeks gestation using linear SVMs are 47\% sensitivity and 57\% specificity showing an improvement of 20\% and 30\% respectively as compared to \citep{Mercer19961885}. 
In addition, we obtain approximately $50\%$ sensitivity and specificity across other data classes and time points. 


We observe that SVM with a non-linear (RBF) kernel performs slightly better than linear SVM for the full data. 
We  believe that a larger data set would highlight the advantage of the non-linear method more prominently. When we consider the entire (full) dataset, the linear/RBF SVM performs better with increasing ticks (T0 to T3 i.e., as the pregnancy progresses).  This reflects our intuition that as we increasingly obtain more information (features) about each patient (example), we expect to better discriminate between them.
SVMs for spontaneous and nulliparous data can sometimes lead to a poor performance. For instance the g-mean for polynomial SVMs with degree 3 for T3 is 0.11 $\pm$ 0.1. 
We consider the nulliparous data only to be the most difficult of the  three datasets.  This is especially clear at T0 when most of the critical features are derived from  previous pregnancy history which is not available for nulliparous women.   The high number of support vectors (not shown in the tables)  required for the SVMs solution throughout  the SVMs runs (across ticks, kernels, data) indicates that the preferable decision rule is approximately linear. In other words, under-fitting the data (small C value) generalizes better to unseen examples.  
We observe a poor performance of the Random Forest method, probably due to overfitting of decision trees on this kind of data. 
We have also shown that the {\em machine-picked} linear model presented in this paper outperforms Creasy table \citep{creasy1980} {\em hand-picked} model. 
In summary, our study demonstrates that model selection and non linear kernels  are promising approaches for prediction of PTB.

\section{Significance and Impact}
\label{future}
%
%
Preterm birth is a challenging and complex real world problem that pushes the boundary of machine learning state-of-the-art methodologies.  
Today, there does not exist an effective prediction system to identify women at risk of PTB to prevent this adverse pregnancy outcome.
Specifically, nulliparous women (first time mothers-to-be) remain the most vulnerable population.

We present a comprehensible and reproducible study that demonstrates that more accurate prediction of preterm birth is not an elusive task. Our best performing algorithms attained (balanced) accuracy rates of 60\%. 
We have demonstrated significant improvement compared to previous prediction performance on the same type of data and developed models that integrate heterogenous risk factors. 



For future work, we plan to conduct larger scale experiments on other sources of data to study preterm birth including existing datasets and other 
data collected from electronic health records at a large urban hospital. 






  \section*{Acknowledgment}
We  thank  the  reviewers  for their helpful  comments. This work is partially funded by the National Science Foundation (NSF) under agreement number IIS-1454855 and IIS-1454814. Any  opinions, findings and  conclusions or recommendations   expressed in this material are those of the author(s) and do not necessarily reflect those of the National Science Foundation.

We also acknowledge the assistance of NICHD, the MFMU Network,  and the Protocol Subcommittee in making the database available 
on behalf of the project. 
The contents of this paper represent the views of the authors and do not 
represent the views of the Eunice Kennedy Shriver National Institute of 
Child Health and Human Development Maternal-Fetal Medicine Units Network 
or the National Institutes of Health.

The authors are very thankful to Dr. Tara Randis and Dr. Mary McCord for their invaluable contributions on preterm birth background and insightful discussions in an early stage of this project. 

\newpage


\bibliographystyle{natbib}
\bibliography{references_anita,ML1,references,references_tara,refML}

\newpage

\section*{Appendix}

Table \ref{table:creasy} shows the Risk of Preterm Delivery system  (RPD) \cite{creasy1980}. This scoring system is a modification of the CPDR system proposed in \cite{papiernik-berkhauer1969}.
The final score is computed by addition of the number of points given any item. 
A final score between 0 and 5 is classified as {\em low risk}; a score between  6 and 9 as {\em medium risk} and any score higher or equal to 10 deemed as {\em high risk} score for preterm birth.

\begin{table*}[htbp]
\begin{center}
\caption{Risk of Preterm Delivery (RPD) \cite{creasy1980}}
{ \scriptsize \begin{tabular}{@{}cp{3.2cm}p{3cm}p{2.2cm}p{3cm}@{}}
\hline 
{\bf Points} & {\bf Socioeconomic status} & {\bf Past history} & {\bf Daily habits} & {\bf Current pregnancy} \\ \hline 
{\bf 1} & 2 children at home & 1 abortion  & Work outside home & Unusual fatigue \\
   & Low socioeconomic status  & Less than 1 year since last birth  & & \\
{\bf 2} & Younger than 20 years Older than 40 years  & 2 abortions  & More than 10 cigarettes per day & Less than 13 kg gain by 32 weeks' gestation \\
& Single parent & & & Albuminuria  \\
& & & & Hypertension\\
& & & & Bacteriuria\\    
{\bf 3} & Very low socioeconomic & 3 abortions & Heavy work & Breech at 32 weeks\\
& status & & Long tiring trip & Weight loss of 2 kg  \\
& Shorter than 150 cm & & & Head engaged \\
& Lighter than 45 kg & & & Febrile illness \\
{\bf 4} & Younger than 18 years	& Pyelonephritis  & &  Metrorrhagia after 12 weeks' gestation\\
& & & & Effacement\\
& & & & Dilatation\\
& & & & Uterine irritability\\
{\bf 5} & &  Uterine anomaly & & Placenta previa \\
& &Second trimester abortion &  & Hydramnios \\ 
& & DES exposure & & \\
{\bf 10} & & Premature delivery & & Twins \\
& &  Repeated second-trimester abortion & & Abdominal surgery \\
\hline
\end{tabular}
}
\medskip
\label{table:creasy}
\end{center}
\end{table*}

Table \ref{RPDmapping1}  shows the mapping we have developed of RPD factors to MFMU features.

\begin{table*}[htbp]
{\scriptsize
\begin{center}
\caption{Mapping of RPD factors to MFMU dataset }
\label{RPDmapping1}
{
 \begin{tabular}{| p{5cm} |l |   l}
 \noalign{\smallskip}
 \hline
   \bf{RPD Factor} & \bf{MFMU Feature} \\ 
 \hline
     2 children at home \hspace*{1cm}			& BPKIDS $\geq$ 2 \\
     Low socioeconomic status \hspace*{1cm}			&  BPINCOME == 1  and SCHOOLYR == (13 or 14 )  \\
     Younger than 20 years \hspace*{1cm}			& AGEMOM $ < 20$ and $\neq 18$ \\
     Older than 40 years\hspace*{1cm}			& AGEMOM $== 40$ \\
     Single parent \hspace*{1cm}			& BPMARITL\_2 $== 1$ or BPMARITL\_3 $== 1$  \\
     Very low socioeconomic status \hspace*{1cm}			& BPPHONE $== 0$ and BPCAR $== 0$ and BPINCOME $== 1$\\& and BPWORK $== 0$ and SCHOOLYR $< 13$ \\
     Shorter than 150 cm \hspace*{1cm}			& HEIGHT $< 59$ \\
     Lighter than 45 kg \hspace*{1cm}			& WGTPRE $< 45$ \\
     Younger than 18 years \hspace*{1cm}			& AGEMOM $= 18$ \\
     1 abortion \hspace*{1cm}			& BPINDUCE $== 1$ \\
     Less than 1 year since last birth \hspace*{1cm}			& LASTPREG $== 0$ \\
     2 abortions \hspace*{1cm}			& BPINDUCE $== 2$ \\
     3 abortions \hspace*{1cm}			& BPINDUCE $> 2$ \\
     Pyelonephritis \hspace*{1cm}			& BPINFEC $== 1$ and PYELO $== 1$\\
     Uterine anomaly \hspace*{1cm}			& BPFIBR $== 1$ or BPLOWER\_2 $== 1$\\
     Second trimester abortion \hspace*{1cm}			& SECAB $> 0$ \\
     DES exposure \hspace*{1cm}			& N/A \\
     Premature delivery \hspace*{1cm}			& PRETERM $== 1$ \\
     Repeated second-trimester abortion \hspace*{1cm}			& N/A \\
     Work outside home \hspace*{1cm}			& BPJOB $== 1$ \\
     More than 10 cigarettes per day \hspace*{1cm}			& BPSMOKE $== 1$ and CIGSPRE $>= 10$ \\
     Heavy work \hspace*{1cm}			& N/A \\
     Long tiring trip \hspace*{1cm}			& N/A \\
          Unusual fatigue \hspace*{1cm}			& BPSTAND $== 1$ or BPBREAK $== 0$ \\& or BPVIBES $== 1$ or BPHRS $> 50$ \\ 
     Less than 13 kg gain by 32 weeks gestation \hspace*{1cm}			& WEIGHTV3 - WEIGHTV1 $< 13$ \\
     Albuminuria \hspace*{1cm}			& BPURINE $> 0$ \\     
     Hypertension \hspace*{1cm}			& BPHYPER $== 1$ \\
%
%
%
     Bacteriuria \hspace*{1cm}			& BACTER $== 1$ \\
     Breech at 32 weeks \hspace*{1cm}			& N/A \\
     Weight loss of 2 kg \hspace*{1cm}			& WEIGHTV3 - WEIGHTV1 $< -2$ \\
     Head engaged \hspace*{1cm}			& N/A \\
     Febrile illness \hspace*{1cm}			& HERPES $== 1$ or VHERPES $== 1$ or CYS $== 1$\\ & or VCYS $== 1$\\
     Metrorrhagia after 12 weeks' gestation \hspace*{1cm}			& BPVAG2ND $== 1$ or PERBLD $== 1$ \\
     Effacement \hspace*{1cm}			& N/A \\
     Dilatation \hspace*{1cm}			& BPCRVLT $ < 25$  \\
     Uterine irritability \hspace*{1cm}			& N/A \\
     Placenta previa \hspace*{1cm}			& N/A \\
     Hydramnios \hspace*{1cm}			& OLIGO $== 1$ \\
     Twins \hspace*{1cm}			& N/A \\
     Abdominal surgery \hspace*{1cm}			& BPABD $== 1$ \\
  \noalign{}\hline
  \end{tabular} }
  \end{center}
  \vspace*{-0.3cm}
}
\end{table*}

\end{document}